\documentclass[11pt]{article} 
\usepackage{rldmsubmit,palatino}
\usepackage{graphicx}
\usepackage{algorithmic}
\usepackage{amsmath,amssymb,amsfonts}
\usepackage{wrapfig}

\title{Autonomous Open-Ended Learning \\ of Tasks with Non-Stationary Interdependencies}

\author{
Alejandro Romero $^{1}$, Gianluca Baldassarre $^{2}$, Richard J. Duro$^{1}$, Vieri Giuliano Santucci$^{2}$\\
$^1$Integrated Group for Engineering Research (GII)\\
CITIC research center\\
Universidade da Coruña, Spain\\
\texttt{\{alejandro.romero.montero, richard.duro\}@udc.es}\\
$^2$Istituto di Scienze
e Tecnologie della Cognizione (ISTC)\\
Consiglio Nazionale delle Ricerche (CNR), Roma, Italy\\
\texttt{\{gianluca.baldassarre, vieri.santucci\}@istc.cnr.it}\\
}

\DeclareMathOperator{\E}{\mathbb{E}}

\begin{document}

\maketitle

\begin{abstract}
Autonomous open-ended learning is a relevant approach in machine learning and robotics, allowing the design of artificial agents able to acquire goals and motor skills without the necessity of user assigned tasks. A crucial issue for this approach is to develop strategies to ensure that agents can maximise their competence on as many tasks as possible in the shortest possible time. Intrinsic motivations have proven to generate a task-agnostic signal to properly allocate the training time amongst goals. While the majority of works in the field of intrinsically motivated open-ended learning focus on scenarios where goals are independent from each other, only few of them studied the autonomous acquisition of interdependent tasks, and even fewer tackled scenarios where goals involve non-stationary interdependencies. Building on previous works, we tackle these crucial issues at the level of decision making (i.e., building strategies to properly select between goals), and we propose a hierarchical architecture that treating sub-tasks selection as a Markov Decision Process is able to properly learn interdependent skills on the basis of intrinsically generated motivations. In particular, we first deepen the analysis of a previous system, showing the importance of incorporating information about the relationships between tasks at a higher level of the architecture (that of goal selection). Then we introduce H-GRAIL, a new system that extends the previous one by adding a new learning layer to store the autonomously acquired sequences of tasks to be able to modify them in case the interdependencies are non-stationary. All systems are tested in a real robotic scenario, with a Baxter robot performing multiple interdependent reaching tasks.\end{abstract}


\acknowledgements{*This work was partially supported by the the MCIU of Spain/FEDER (grant RTI2018-101114-B-I00), Xunta de Galicia (EDC431C-2021/39), Centro de Investigación de Galicia "CITIC" (ED431G 2019/01), by the Spanish Ministry of
Education, Culture and Sports for the FPU grant of Alejandro Romero, and by the European Union’s Horizon 2020 Research and Innovation Programme under Grant Agreement no 713010, Project “GOAL-Robots – Goal-based Open-ended Autonomous Learning Robots”}

\startmain 

\section{Introduction}

Autonomous open-ended learning (A-OEL) \cite{Colas2020intrinsically,Santucci2020intrinsically} aims at the development of artificial agents able to solve a potentially unbounded set of different tasks in environments that might be unknown at design time. Similarly to multi-task reinforcement learning \cite{Florensa2018}, a system has to learn multiple policies associated with different goals (i.e., the achievement of desired states/effects in the environment. In this sense ``task'' and ``goal'' can be used interchangeably, where a task consists in the achievement of the associated goal). However, in the A-OEL perspective the focus is not ``simply'' on the maximisation of the rewards, but on the development of a strategy that allows the agent to properly allocate the training time to maximise its competence over all the goals during the learning period. This reflects the scenario in which a system is left to explore the world for a limited time, while only in a second phase it will be assigned tasks that are useful for users: the greater the competence acquired, the higher the probability of being able to maximise the rewards for the subsequently assigned tasks. 

In this perspective, intrinsic motivations (IMs) have been used in the field of machine learning and developmental robotics \cite{Oudeyer2007intrinsic,Baldassarre2013Book}, amongst other applications, to provide self-generated signals guiding the autonomous selection of tasks to be trained \cite{Santucci2013best,Baranes2013,blaes2019control}. The majority of works within the intrinsically motivated open-ended learning framework are normally focused on scenarios where goal achievability does not depend on specific environmental states or preconditions. However, in real-world scenarios, tasks may require particular conditions to be fulfilled or, more interestingly, they may be interdependent so that one (or a sequence of them) is the precondition for the achievement of the other(s). As an example, consider a setting where the goal of arriving at a particular $[x,y]$ location is possible only if the intensity of illumination has already been set to a certain value. In that regard, the navigation goal is conditioned on the agent having caused the environment to reach a specific ``illumination intensity goal''.

The ``interdependent tasks'' scenario is of particular interest for both machine learning and robotics, and it has so far been scarcely studied in an A-OEL perspective \cite{Forestier2017,Santucci2019autonomous,blaes2019control}. Under the headings of curriculum learning \cite{Matiisen2019,Narvekar2019} and hierarchical reinforcement learning \cite{Bakker2004,Niel2018}, different works have focused on sequencing ever more complex tasks, with the aim of transferring knowledge from one to another or dividing the most difficult goals into sub-goals that can be learnt more easily. However, in most of these works, even when the agent autonomously creates and selects sub-tasks, these processes are based on an externally-assigned final goal. On the contrary, here we are interested in a situation where several possible interrelated tasks are presented to the agent, whose aim is to maximise its overall competence by selecting the goals it wants to learn and, where necessary, to learn the different curricula that are needed to acquire the skills of the hierarchically more complex goals. Furthermore, our aim is to address an even more complex scenario, which to date, especially in the field of autonomous robotics, is still poorly addressed: the scenario in which the interdependencies between goals may change over time, thus forcing the system to re-learn the different sequences in order to maintain a high competence in solving all the possible tasks.

In previous works, we presented the M-GRAIL architecture \cite{Santucci2019autonomous,Romero2021} and we analysed how treating goal selection as a Markov Decision Process (MDP) results in better overall competence acquisition with respect to other approaches that treat goal selection as a bandit or contextual bandit problem \cite{Santucci2016,Forestier2017,blaes2019control}. Here we present a twofold study. On the one hand, we deepen the analysis of our approach by comparing M-GRAIL with a system which, although treating task selection as a bandit problem, is able to integrate the information about the dependencies between the goals directly into the low-level skills. On the other hand, we extend our system which, by relying only on intrinsic motivations, was not able to save the acquired sequences, and we present H-GRAIL, a new hierarchical system with 3 different learning processes. H-GRAIL is then tested in a scenario where the interdependencies between goals can change over time.   

\section{Problem Analysis and Suggested Solution}

In multiple task learning the objective is to learn a set of different tasks (i.e., reaching a different goal), each associated to a core MDP. We assume a goal $g$ is a specific subset $S^g \in S$, so that $g$ has been achieved if the system enters any state in $S^g$. For each goal $g$ there is a goal-dependent reward function $r^g$, determining a goal-dependent reward $r^g_t$ at time $t$. Following \cite{Florensa2018}, the overall objective of the system is then to find a policy $\pi$ (or different goal-related policies $\pi^g$) such that
\begin{equation}\label{eqn:objFunc1}
   \pi^* = \underset{\pi}{\text{argmax}} \E_{g\sim P} \big[ r^g(\pi) \big],
\end{equation}
where $P$ is a probability distribution over the set $G$ of possible goals $g$. As analysed in \cite{Santucci2019autonomous}, in an OEL scenario the system objective is to maximise, in a finite and unknown learning time $L$, its competence $C$ over $G$
\begin{equation}\label{eqn:objfuncCompetence}
    C = \E_{g \sim P} C^g
\end{equation}
where $C^g$ is the goal-related competence for achieving goal $g$ using $\pi^g$. In this sense, $C^g$ reflects the expected goal-specific rewards $\E\{r^g_t + r^g_{t+1} + \ldots |\pi^g\}$ when executing $\pi^g$. Since $L$ is finite and unknown, the agent has to maximise $C$ as quickly as possible, efficiently distributing the learning time over $G$. The A-OEL of multiple goals is thus a training time allocation problem where the system has to build a meta-policy $\Pi$ that at each time step $t$ selects a goal to train for a certain (eventually fixed) amount of time $t'$ so that at $L$ the overall competence $C$ will be maximal.

\begin{equation}\label{eqn:objFunc2}
   \Pi^*(t) =  \underset{\Pi}{\arg\max} \,\, \E_{g\sim P} \big[ C^g(t+L) \,|\, \Pi (t) \big].
\end{equation}

Given this objective, $\Pi$ is not selecting goals to train with respect to their current competence $C^g(t)$, i.e. to the amount of returns expected for executing $\pi^g$, but with respect to the amount of competence the system can gain for practicing on $g$. The reward is thus the intrinsic motivation signal determined by the competence improvement $\Delta C^g = C^g(t+t') - C^g(t)$ obtained for training on $g$. Autonomously learning multiple tasks can thus be seen as a two-level problem: \textit{(a)} the high-level goal-selection process to increase competence; and \textit{(b)} the low-level learning of policies $\pi^g$. While we make no assumption about which algorithm is used to solve \textit{(b)}, in the case where goals are independent (learning about one goal does not help the agent
achieve other goals) and where the initial state of the environment is not affecting policy execution, task selection can be modelled as an $N$-armed bandit, as it has been typically addressed in the majority of OEL architectures, e.g. \cite{Baranes2013,Santucci2016,blaes2019control}. Differently, if tasks are interdependent (i.e. the achievement of some goals constitute the precondition for the achievement of other goals), $\Pi$ has to take into consideration that goal-selection implies long-term consequences in terms of possible future rewards (i.e. future competence gain): it may be important to spend time practising goals that, on their own, do not provide competence improvement, but that allow the agent to train more advantageous goals. Thus, task selection now involves a credit assignment problem, since evaluating the returns gained from goal $g$ requires backing up expected future competence improvement that may be achieved by further practising ``distant'' goals that have $g$ as a precondition.

For this reason, similarly to \cite{Narvekar2019}, in M-GRAIL we proposed to no longer treat goal selection as a bandit problem but as an MDP, and to solve it through a Q-Learning algorithm that models the relation between the values of interrelated goals. However, given the transient nature of intrinsic motivations (which disappear when competence has reached a plateau), M-GRAIL has the problem that although it is able to learn all the policies $\pi^g$ by properly assigning value to goals that constitute preconditions for others, it is not able to execute these sequences once the evaluations determined by intrinsic motivations have vanished. To cope with this limitation, here we propose H-GRAIL, a new architecture that adds a further layer to the previous system. M-GRAIL (as well as other systems in the literature) was essentially composed of two levels: one in which low level skills are learnt, through the maximisation of goal-specific rewards $r^g$; and a higher level in which goal selection takes place, treated as an MDP in which a Q-Learning algorithm maximises the competence improvement signal $\Delta C^g$. H-GRAIL adds a further layer, changing the structure of the goal selector and dividing it into two components. The meta policy $\Pi$ is again treated as a bandit problem based on the maximisation of competence improvement, but once the system has selected a goal $g$, it is used as an input to a second level which, structured as an MDP, must learn to sequence the sub-goals necessary to achieve the task selected by $\Pi$. This sub-goal selector, unlike the M-GRAIL selector, aims at the maximisation of goal-specific reinforcement $r^g$: in this way, the learnt interdependencies between the different goals will remain available to the system as ``curricula'', or better, as policies over sub-goals, even after the intrinsic motivations have disappeared. 

\section{Robotic Setup and Experiments}
\label{sec:M-GRAIL}

\begin{wrapfigure}{r}{0.35\textwidth}
    \centering
    \includegraphics[width=4.3cm]{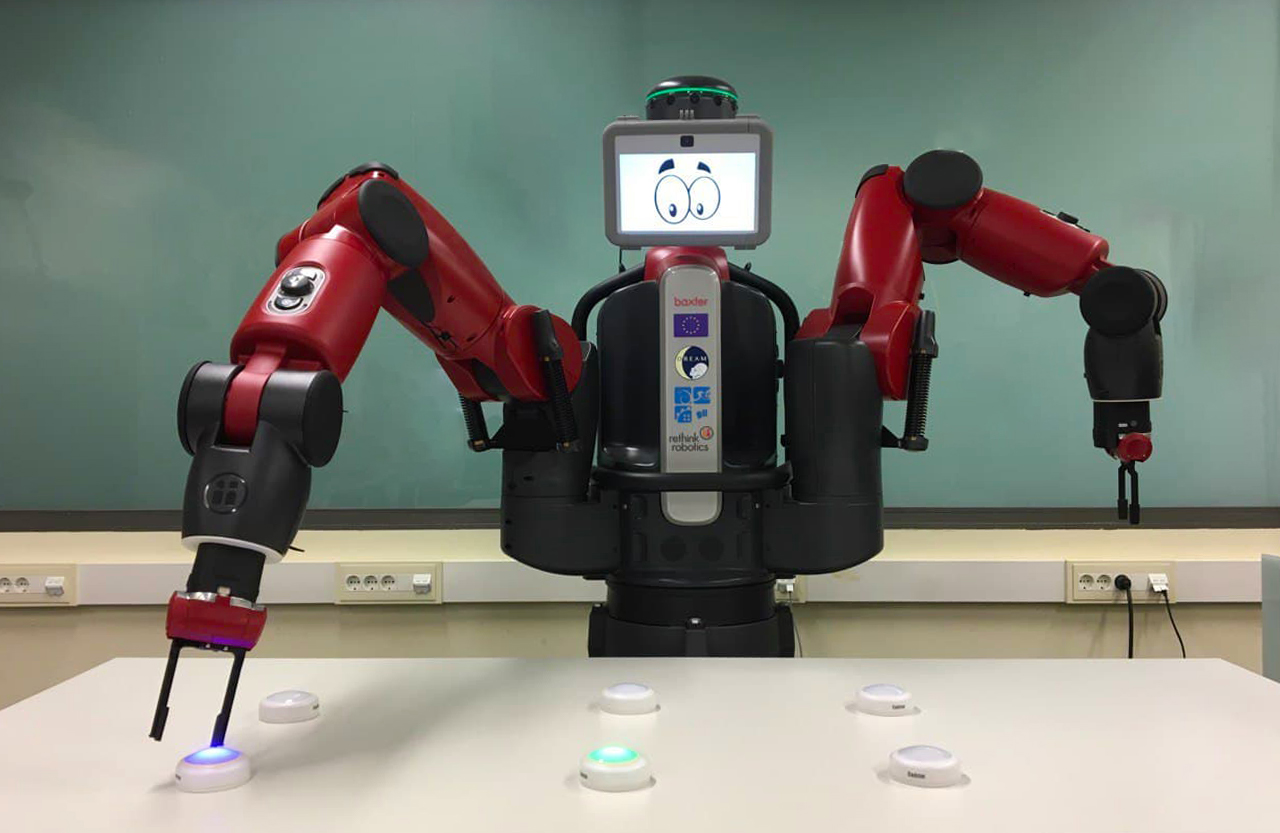}
    \caption{The experimental setup: buttons ``light up'' when pressed if all their preconditions are satisfied.}
    \label{fig:Setup}
\end{wrapfigure}

To test our system we implemented a robotic scenario (Fig. \ref{fig:Setup}) where a Baxter robot has to learn to reach for different buttons that ``light up'' when pressed if their preconditions are satisfied. In a first experiment (Sec. \ref{sec:MGRAILMDB}), we compare M-GRAIL with a modified version of the e-MDB system \cite{Romero2020}, called Bandit-MDB, where the motivational system is implemented using a goal-selecting bandit mechanism based on competence improvement intrinsic reinforcements. On the contrary, M-GRAIL treats goal-selection as an MDP and solves it through a standard Q-Learning algorithm \cite{Watkins1992}. Both M-GRAIL and Bandit-MDB learn low-level skills via utility models, where each skill is an artificial neural network-based value function (see \cite{Romero2019} for more details), however only the Bandit-MDB utility models receive contextual information as input (here a binary vector stating if a goal has been achieved within the current epoch, i.e. if a button is ``on'' or ``off''), while in the case of M-GRAIL only the goal selector receives such information. This allow us to analyse how, and in particular at what level of the architecture, a robotic system should handle the dependencies between goals.

In a second experiment (Sec. \ref{sec:Res_HGRAIL}) we test H-GRAIL in a similar robotic scenario, where interdependencies between goals are non-stationary (in particular, they change after a certain time during learning). H-GRAIL receives contextual information regarding the goals at the level of the sub-goal selector (implemented as a Q-Learning algorithm), while the high-level goal selector is implemented as a standard bandit maximising competence improvement.

In both experiments we use the right arm of the robot and we control its wrist final position (x, y, z) through Cartesian position control. Regarding perception, we used the images from an RGB-D camera located on the ceiling of the room and binary sensors associated with buttons being pushed. This information is re-described in the form of distances between the detected objects and the end effector of the robot arm. Therefore, the perception of the robot is $(d_1,\ldots,d_n, s_1 ,\ldots,s_n)$, where $d_j$ are the relative distances between the buttons and the robot end-effector, $s_i$ are the states of the different buttons (active or not), and $n$ is the number of buttons in the scenario (6 in the current experiment).

\begin{figure}[t!]
    \begin{center}
    \includegraphics[width=.70 \textwidth,keepaspectratio]{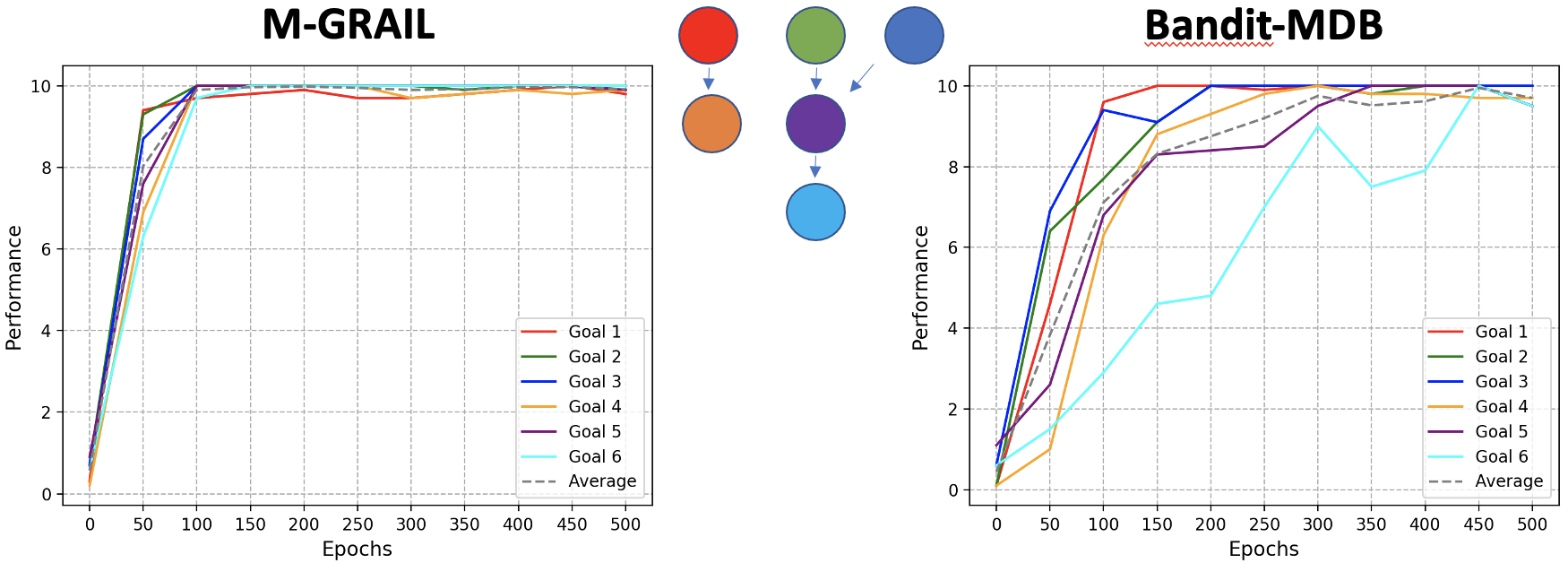}
    \end{center}
    \caption{Configuration of the first experiment (arrows indicate dependencies) and the performance of the two systems.}
    \label{fig:Exp1}
\end{figure}

%

\section{Results and Discussion}
\subsection{Testing where to incorporate interdependence knowledge}
\label{sec:MGRAILMDB}

The first experiment was run for 500 epochs, each lasting 8 trials ending when the robot lights up the target button (it achieves the selected goal) or after a timeout of 70 time steps. At each trial the goal selector of the systems selects the goal to be achieved, while at the end of each epoch we reset the environment (the robot is set to home positions and the buttons are switched off). In this scenario there are two chains of dependencies: a simple one (with just one precondition) and a more complex chain where reaching the last goal (cyan button) requires the accomplishment of three other precondition goals. Fig. \ref{fig:Exp1} shows the performance of the two systems (averages over 20 repetitions). M-GRAIL is very efficient in learning all the tasks, including those requiring preconditions, while Bandit-MDB needs 300 epochs to reach 90\% performance and has more trouble learning the last task (activation of the cyan button). The longer time required by the Bandit-MDB system is the result of it having to learn more complex and longer skills, since, for example, to reach the blue button, the utility model corresponding to that skill must first learn to reach the red and green buttons. Unlike Bandit-MDB, when learning simple skills and concatenating them, M-GRAIL only has to learn to reach the blue button while the goal selector takes care of selecting the precondition goals before it: this ensures that the red and green buttons are active when the blue-button goal is selected. This experiment further corroborates the findings in \cite{Santucci2019autonomous}, showing how our proposal to treat autonomous goal learning in a hierarchical manner yields positive results. In particular, here we analysed how it is more efficient to learn simple skills, not storing information on goal interdependencies, and concatenate goals at the level of the goal selector. This is because the time required for learning a single skill, given the appropriate preconditions guaranteed by the goal selection, is less than the time necessary to learn a skill to achieve a goal and also all those being the preconditions to it. 

\begin{figure}
    \begin{center}
    \includegraphics[width=.65 \textwidth,keepaspectratio]{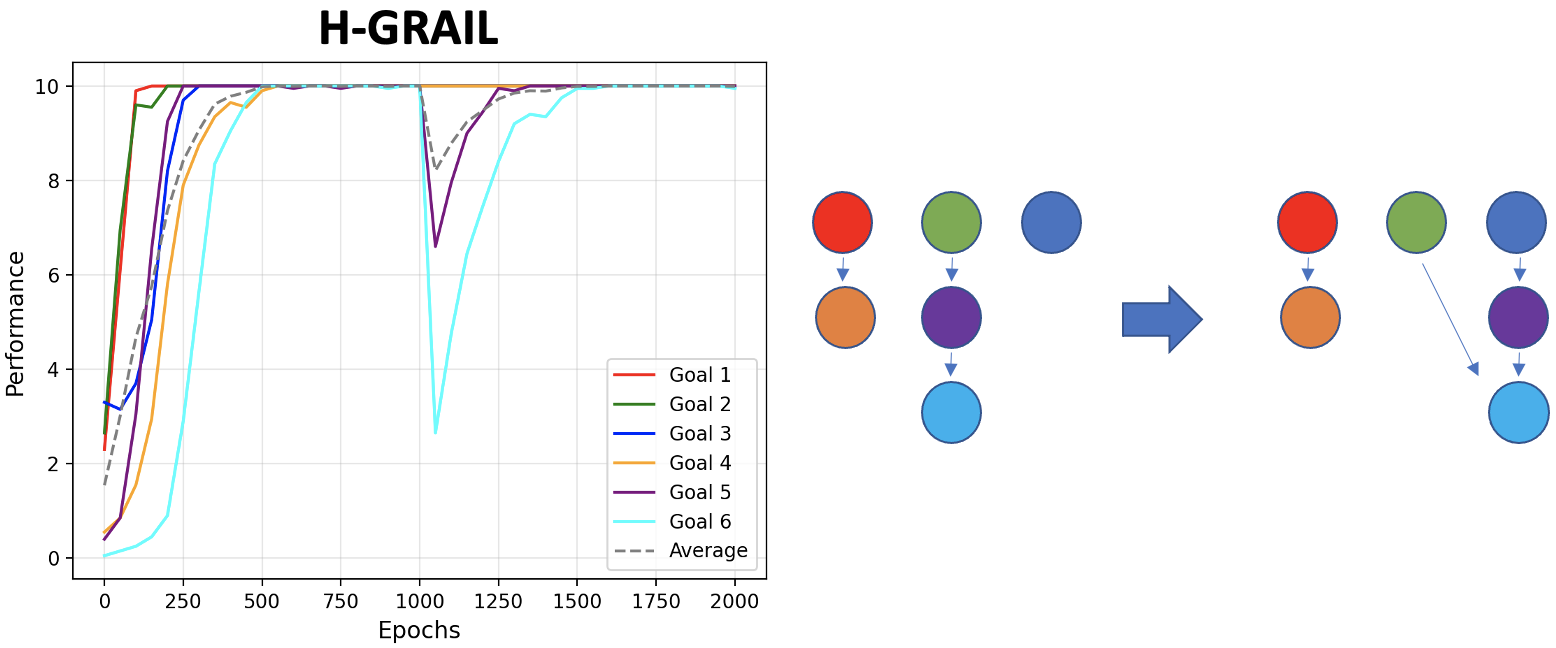}
    \end{center}
    \caption{Configuration of the second experiment with non-stationary dependencies, and the performance of H-GRAIL.}
    \label{fig:Exp2}
\end{figure}

\subsection{Testing H-GRAIL on tasks with non-stationary interdependencies}
\label{sec:Res_HGRAIL}

In the second experiment we test our new architecture H-GRAIL in an environment where interdependencies between goals are non-stationary. In particular, after 1,000 epochs the dependencies are modified as shown in Fig. \ref{fig:Exp2} (right). Experiments are run for a total of 2,000 epochs, composed as in the first experiment. The results (averages over 20 repetitions) show that H-GRAIL was not only able to learn the skills necessary to achieve the different goal-dependent tasks, but also had the ability to dynamically adapt to a change in the structure of task relationships. Unlike M-GRAIL, the sub-goal selector added in H-GRAIL is able to store the acquired curricula through a Q-Learning algorithm based on the reinforcements obtained to achieve the tasks selected by the same sub-goal selector. However, when the interdependencies change the performance in achieving goals will drop (the curricula will not work anymore), thus the goal selector starts again a selection process aimed at maximising competence improvement that pushes the sub-goal selector towards finding new ways to sequence the different tasks to properly achieve those selected by the agent.

\bibliographystyle{abbrv}
\bibliography{RLDMBib}

\end{document}